\documentclass[10pt,twocolumn,letterpaper]{article}

\usepackage{wacv}

\usepackage{enumitem}

\newcommand{\safeincludegraphics}[2][]{%
  \IfFileExists{#2}{\includegraphics[#1]{#2}}{%
    \fbox{\parbox[c][0.14\textheight][c]{0.94\linewidth}{%
      \centering Missing figure: \texttt{#2}}}}}

\usepackage{booktabs}
\usepackage{caption}
\usepackage{microtype}
\usepackage{subcaption}
\usepackage{threeparttable}

\usepackage{booktabs}       
\usepackage{caption}        
\usepackage{enumitem}       

\usepackage{caption}

\usepackage{enumitem}

\definecolor{wacvblue}{rgb}{0.21,0.49,0.74}
\usepackage[breaklinks,colorlinks,allcolors=wacvblue]{hyperref}

\title{Auditing Observation-Quality Shortcuts in the Evaluation of Person-Centered Motion for Video Anomaly Detection}

\author{
Chaehyun Kim \qquad Changwon Lim$^{*}$\\
Department of Statistics and Data Science\\
Chung-Ang University\\
{\tt\small \{rea828, clim\}@cau.ac.kr}\\
{\small $^{*}$Corresponding author}
}

\hypersetup{
  pdftitle={Are We Detecting Abnormal Motion? Auditing Observation-Quality Shortcuts in Person-Centered Video Anomaly Detection},
  pdfauthor={Chaehyun Kim and Changwon Lim}
}

\begin{document}
\maketitle

\begin{abstract}
Video anomaly detection (VAD) identifies deviations from normal patterns, but such deviations do not necessarily correspond to safety-critical events. We focus on human-centered emergencies and violence and investigate whether motion-based anomaly discrimination reflects task-relevant motion information or observation-quality differences arising from person detection, pose estimation, and tracking. We introduce an observation-quality-aware evaluation framework that combines quality diagnostics, quality balancing, motion orthogonalization, and robustness analyses. On human-centered categories of UCF-Crime, quality-only features match or exceed the discrimination of multi-cue motion in the original cohort, revealing a strong observation-quality shortcut. After cardinality matching brings all 11 measured quality variables below a prespecified marginal balance threshold, multi-cue motion consistently outperforms quality-only features across Logistic Regression, Random Forest, and RBF-SVM, yielding ROC-AUC gains of 0.046-0.141. Because nonlinear quality-only prediction remains above chance after matching, we interpret this result as additional discrimination beyond measured marginal quality rather than as evidence of quality-free or causal motion understanding. These findings show that high VAD performance should not be interpreted as evidence of abnormal-motion modeling without auditing the reliability of the person-observation pipeline. Our study provides a systematic protocol for evaluating observation-quality shortcuts and for assessing whether motion-based discrimination survives explicit control of detection, pose-estimation, and tracking reliability.
\end{abstract}

\section{Introduction}
\label{sec:intro}

Video anomaly detection (VAD) is designed to distinguish events that depart from normal patterns, but general VAD benchmarks often group together abnormal events with substantially different visual manifestations. UCF-Crime, for example, introduced long, untrimmed surveillance videos with video-level labels for 13 anomaly categories, ranging from human violence to road accidents, explosions, and property-related events~\cite{sultani2018}. Such breadth is useful for general anomaly detection, but not every anomaly is defined primarily by how people move. This distinction matters when evaluating person-centered VAD methods, which model human trajectories or skeleton dynamics to reduce reliance on background appearance~\cite{morais2019,markovitz2020,hirschorn2023}. To make motion-based discrimination interpretable with respect to task-relevant human behavior, we therefore focus on human-centered safety-critical categories--assault, abuse, fighting, and shooting--in which human actions and interactions provide relevant visual evidence. This restricted scope is not intended to redefine VAD as violence recognition, but to provide a setting in which the contribution of person-centered motion can be meaningfully audited.

\begin{figure*}[t]
    \centering
    \includegraphics[width=\textwidth]{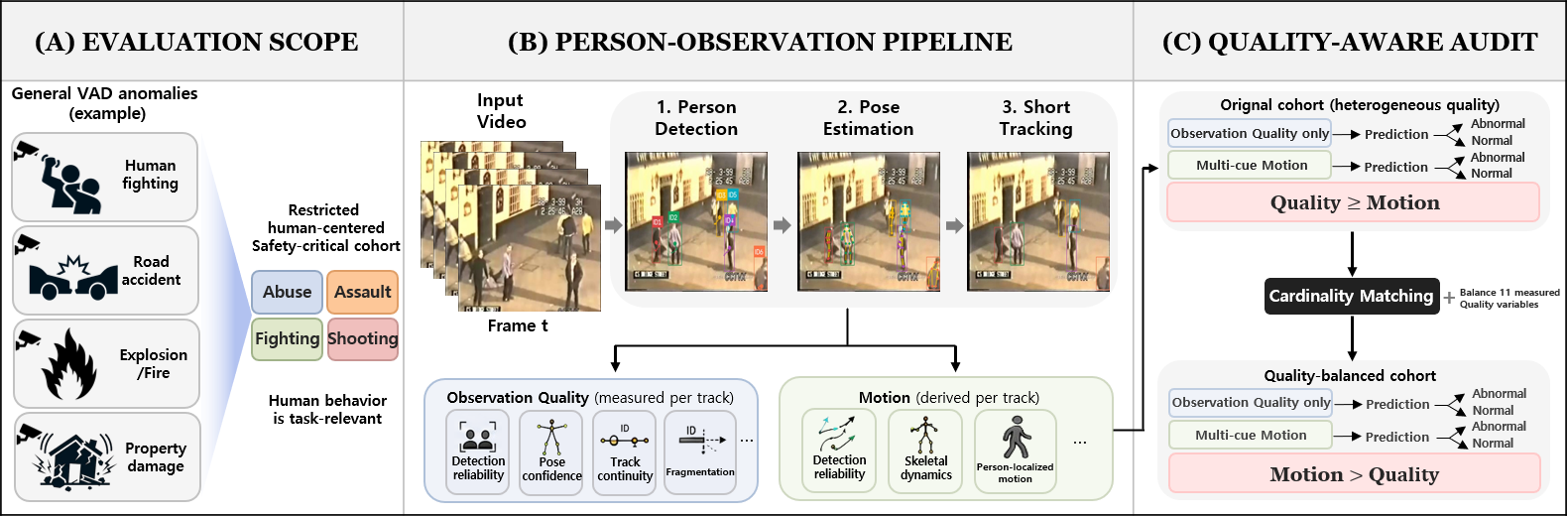}

    \captionsetup{width=\textwidth}
    \caption{\textbf{Overview of the observation-quality-aware evaluation framework.}
    (a) We restrict the heterogeneous UCF-Crime anomaly space to human-centered safety-critical categories in which human behavior is task-relevant. (b) Detection, pose estimation, and tracking provide both person-centered motion and separately measured observation-quality variables. (c) Observation quality is highly discriminative in the original cohort; after cardinality matching balances 11 measured quality variables, multi-cue motion retains greater discrimination than quality alone. The protocol tests whether motion provides additional discrimination after balancing measured observation quality, rather than establishing causal motion effects.}
    \label{fig:overview}
\end{figure*}

The central difficulty is that person-centered motion is not observed directly, but is derived from an imperfect person-observation pipeline (Fig.~\ref{fig:overview}). Missed detections create gaps, uncertain keypoints perturb trajectories, unstable boxes alter person crops, and fragmented tracks introduce discontinuities. Pose-based VAD studies likewise identify noisy skeletons, lost tracks, and missing observations as central practical failure modes~\cite{noghre2023challenges}. These effects are not peripheral implementation details. They determine the observations from which the motion representation itself is computed.

This leads to a direct evaluation objective: quantify how much of person-centered VAD discrimination is already present in the reliability of the observation pipeline. We call label-predictive detection, pose, and tracking reliability an \emph{observation-quality shortcut}. In this paper, a shortcut is a task-adjacent benchmark signal that predicts the label without an explicit description of human motion, consistent with the broader literature on shortcut learning and spurious predictive cues~\cite{geirhos2020shortcut,hermann2023foundations,ye2024cleverhans}. When this signal is strong, the usual inference
\begin{equation}
\substack{\text{\small High VAD}\\[-1pt]\text{\small performance}}
\;\not\Rightarrow\;
\substack{\text{\small evidence of abnormal-}\\[-1pt]\text{\small motion modeling}}
\label{eq:not_imply}
\end{equation}

does not hold without an additional audit. Importantly, we treat observation quality as a competing predictive signal, not as a causal confound whose removal identifies a causal motion effect.
We demonstrate that this problem is substantial on human-centered categories of UCF-Crime. In the original eligible cohort, observation quality alone reaches ROC-AUC 0.827--0.862 across three classifier families. The corresponding multi-cue motion representation reaches 0.821--0.844. Thus, before any quality control, the reliability of person observation is at least as discriminative as the motion representation that is intended to describe behavior. Tracking continuity and fragmentation account for most of this effect.

The stronger objective is to determine whether motion remains discriminative after abnormal and normal videos are balanced on measured observation quality. We construct a quality-balanced cohort using cardinality matching, which is designed to maximize sample retention subject to explicit balance constraints~\cite{zubizarreta2012,niknam2022cardinality,visconti2018limited}, with a prespecified feature-wise balance requirement. Observation quality alone matches or exceeds multi-cue motion across all three classifier families in the original cohort. After measured-quality balancing, this ordering reverses consistently for the combined motion representation.

The original UCF-Crime cohort contains a strong observation-quality shortcut, so unadjusted motion performance has an ambiguous interpretation. After balancing the measured marginal quality variables, multi-cue motion retains additional discrimination across all three classifier families. Quality-only RF remains predictive and category effects differ; matching and the present handcrafted representation therefore establish neither semantic understanding nor generalization beyond this dataset and fixed pipeline.

Our contribution is an evaluation protocol rather than a new VAD architecture. Following the structure of modern benchmark and dataset-evaluation studies, we first identify a concrete limitation of the existing evaluation setting, then construct a controlled protocol that isolates the limitation, and finally quantify how conclusions change under that protocol. Specifically:
\begin{itemize}[
    leftmargin=*,
    topsep=1pt,
    itemsep=1pt,
    parsep=0pt,
    partopsep=0pt
]
    \item \textbf{Observation-quality audit.} We define an 11-variable detection--pose--tracking quality vector and show that observation reliability alone can match or exceed multi-cue motion in the original cohort.
    \item \textbf{Measured-quality-balanced evaluation.} We use cardinality matching to construct a cohort satisfying a prespecified feature-wise marginal-balance criterion and re-evaluate motion against the remaining quality-only baseline.
    \item \textbf{A robust but bounded change in evaluation conclusion.} After matching, multi-cue motion exceeds quality-only prediction across three classifier families, with the same direction preserved under pair-aware validation and fold-wise orthogonalization; temporal and category analyses further reveal where this conclusion does and does not hold.
\end{itemize}
The key message is therefore not that person-centered motion is invalid. It is that claims about person-centered motion require an observation-quality audit. The proposed protocol makes that requirement measurable.

\section{Related Work}
\label{sec:related}

\subsection{Video Anomaly Detection and Evaluation}

VAD has been studied through reconstruction, prediction, multiple-instance learning, and pretrained spatiotemporal representations. UCF-Crime~\cite{sultani2018} provides 1,900 long, untrimmed surveillance videos with video-level labels spanning 13 heterogeneous anomaly categories. This breadth enables broad anomaly evaluation, but aggregate performance can conflate qualitatively different visual evidence across event types.

Recent work has therefore questioned whether standard VAD benchmarks and metrics faithfully isolate the behavior of interest, including temporal localization, early detection, scene dependence, and other sources of evaluation bias~\cite{liu2025rethinking}. Our study follows this evaluation-oriented perspective but focuses on a different source of ambiguity: for person-centered VAD, the motion representation itself depends on the reliability of person detection, pose estimation, and tracking. We therefore evaluate whether motion-based discrimination remains informative after explicitly auditing and balancing these measured observation-quality factors.

\subsection{Person-Centered Motion Representations}

Person-centered VAD represents detected people using trajectories, pose graphs, or localized motion~\cite{morais2019,markovitz2020,hirschorn2023}. Learned skeleton systems include latent-space contraction in COSKAD~\cite{flaborea2024coskad} and confidence-weighted keypoint density estimation in SeeKer~\cite{delic2025seeker}. Pose-based VAD studies identify noisy skeletons, lost tracks, and missing information as central failure modes~\cite{noghre2023challenges}.

Their common dependency is the person-observation pipeline. Pelvis trajectories require reliable keypoints and temporal association; box-center trajectories require stable detections and tracking; person-localized flow depends on stable crops; and even motion summaries computed from valid tracks inherit missing observations and fragmentation. The observation process therefore changes the measured motion whenever detection, pose, or tracking fails~\cite{noghre2023challenges}.

\subsection{Dataset Shortcuts and Evaluation Audits}

Shortcut learning describes predictive rules that succeed on a benchmark without capturing the intended mechanism~\cite{geirhos2020shortcut,hermann2023foundations}, a phenomenon documented across modalities under related framings such as spurious correlations and Clever Hans behavior~\cite{ye2024cleverhans}. In video, static appearance substitutes for temporal reasoning~\cite{li2023static,lei2022singleframe,kowal2022deeper,dave2024noshortcuts}; VAD rankings likewise depend on scene and metric construction~\cite{liu2025rethinking}. Audits therefore need competing baselines, controlled cohorts, and stress tests rather than aggregate scores alone.

Most closely related to our work, Abdulaziz and Bondarev audit UCF-Crime evaluation with respect to metric granularity, resolution, and scene bias~\cite{abdulaziz2026audit}. Our analysis is complementary: rather than auditing how
frame-level metrics or recording properties affect benchmark scores, we audit the reliability of the person-observation pipeline from which person-centered motion is constructed. We explicitly quantify detection--pose--tracking quality-only
predictability, balance the measured quality variables, and re-evaluate motion on a matched cohort.

Our audit tests observation reliability as a competing explanation and balances measured quality before re-evaluating motion. Cardinality matching maximizes retention under explicit constraints~\cite{zubizarreta2012,niknam2022cardinality,visconti2018limited}; importantly, marginal mean balance need not balance interactions or joint distributions~\cite{resa2016matching}. We therefore report both SMDs and post-match quality-only prediction, using matching for evaluation control rather than causal identification.

To our knowledge, person-centered VAD evaluation has not jointly reported (i) detection--pose--tracking quality-only predictability, (ii) explicit feature-wise quality balance, (iii) matched re-evaluation of multiple motion representations, and (iv) robustness tests that determine whether the motion-versus-quality conclusion survives pair grouping, residualization, temporal aggregation, and semantic scope changes.


\section{Problem Setting}
\label{sec:problem}

\subsection{Task}

Let $v\in\mathcal{V}$ denote a video and $y(v)\in\{0,1\}$ its video-level label, with $y=1$ for Abuse, Assault, Fighting, or Shooting and $y=0$ for Normal. A classifier produces a score $f(v)$ for abnormal-vs-Normal discrimination. The primary task is video-level discrimination, not temporal localization.

Each video yields an observation-quality vector $Q(v)$ and a motion representation $M_r(v)$:
\begin{align}
Q(v) &\in \mathbb{R}^{11},\\
M_r(v) &\in \mathbb{R}^{d_r},
\end{align}
where $r$ indexes pelvis motion, bounding-box-center motion, person-localized flow, whole-frame flow, or a concatenated representation.

We define two empirical quantities under the same cohort, classifier family, fold assignment, and metric:
\begin{align}
\mathcal{P}_Q &= \operatorname{Perf}\!\left(f_Q(Q(v)),y(v)\right),\\
\mathcal{P}_{M_r} &= \operatorname{Perf}\!\left(f_r(M_r(v)),y(v)\right),\\
\Delta_r &= \mathcal{P}_{M_r}-\mathcal{P}_Q.
\end{align}
ROC-AUC is the primary instantiation of $\operatorname{Perf}(\cdot)$.

The audit asks three direct questions:
\begin{enumerate}[leftmargin=*]
    \item \textbf{Shortcut strength:} quantify label prediction from $Q$ without explicit motion.
    \item \textbf{Controlled motion advantage:} evaluate whether $\Delta_r$ remains positive after balancing the measured quality variables.
    \item \textbf{Evaluation stability:} measure the sign and magnitude of $\Delta_r$ across classifiers, representations, temporal aggregation, and pair grouping.
\end{enumerate}

\subsection{Interpretation}

The analysis is comparative, not causal. A strong $\mathcal{P}_Q$ establishes that observation reliability is label-predictive and therefore available as a shortcut in the benchmark distribution. It does not identify why that association arose. Likewise, matching controls the measured marginal quality variables according to the stated balance criterion; it does not claim to remove every unmeasured or nonlinear source of observability.

Accordingly, all matched results are interpreted as evidence that motion-derived discrimination survives explicit control of measured marginal observation quality, not as deconfounded or causal estimates of motion information.

This distinction does not weaken the primary evaluation claim. The question is whether standard VAD performance remains interpretable when a strong alternative signal is present. If $Q$ alone separates the classes as well as motion, unadjusted AUC is insufficient evidence of motion modeling. If motion then exceeds $Q$ after explicit quality balancing, the evidence for motion becomes stronger.

\section{Observation-Quality Audit}
\subsection{Quality Vector}

We define observation quality as the reliability with which the fixed detection--pose--tracking pipeline observes people, not as the semantic content of the scene. The initial taxonomy contains 16 candidate variables. Constant and exactly duplicated variables are removed before label prediction or matching, yielding the 11 active features summarized in Table \ref{tab:qv1}:
\begin{equation}
Q(v)=
\left[
Q_{\mathrm{det}}(v),
Q_{\mathrm{pose}}(v),
Q_{\mathrm{track}}(v)
\right]\in\mathbb{R}^{11}.
\end{equation}

\begin{table}[!t]
\centering
\small
\setlength{\tabcolsep}{4pt}
\begin{tabular*}{\columnwidth}{@{}l@{\extracolsep{\fill}}p{0.75\linewidth}@{}}
\toprule
Group & Active variables \\
\midrule
Detection & mean / median confidence; valid-detection ratio \\
Pose & mean / median keypoint confidence; low-conf. ratio \\
Tracking & continuity; fragmentation rate; mean / longest gap; reappearance rate \\
\bottomrule
\end{tabular*}
\caption{The 11 active $Q$ observation-quality variables. Tracking counts and gaps are normalized by exposure or time where appropriate.}
\label{tab:qv1}
\end{table}

The quality vector excludes video duration, frame count, frame rate, crowd density, mean number of persons, raw track counts, and processing speed. These variables describe recording conditions, scene content, exposure, or computation rather than observation reliability itself. This separation ensures that a high quality-only score reflects detection--pose--tracking behavior rather than a hidden mixture of metadata and scene statistics.

A video with no retained person detection has no defined pose or tracking-quality profile and is excluded from the primary $Q$ analysis rather than being assigned an artificial zero vector. The resulting estimand is therefore conditional on videos for which the fixed pipeline produces at least one person observation.

\subsection{Stage 1: Quality-Only Diagnostic}

The first stage asks whether the label is already predictable from the observation process. We fit detection-only, pose-only, tracking-only, and full-$Q$ classifiers using exactly the same repeated cross-validation protocol used for motion. A strong $Q$-only result is the diagnostic signature of an observation-quality shortcut in the sense of a label-predictive benchmark cue that need not encode the intended mechanism~\cite{geirhos2020shortcut,hermann2023foundations}.

\subsection{Stage 2: Measured-Quality Balancing}

We construct 1:1 abnormal--Normal matched cohorts without replacement using only the 11 $Q$ variables. Our target is feature-wise marginal mean balance in measured observation quality, rather than equality of the full joint quality distribution. Matching is used here as an evaluation-control device, following the balance-constrained matching literature~\cite{zubizarreta2012,niknam2022cardinality,visconti2018limited}. Balance is assessed feature-wise by
\begin{equation}
|\mathrm{SMD}_k|=
\left|
\frac{\bar{Q}_{k,1}-\bar{Q}_{k,0}}
{\sqrt{(s_{k,1}^2+s_{k,0}^2)/2}}
\right|.
\end{equation}
The primary acceptance rule is
\begin{equation}
\max_k |\mathrm{SMD}_k| < 0.10.
\label{eq:smdcrit}
\end{equation}

We compare global nearest-neighbor assignment with a caliper against cardinality matching, which explicitly optimizes sample retention subject to balance constraints~\cite{zubizarreta2012,niknam2022cardinality}. The selection rule is fixed: first satisfy Eq.~\ref{eq:smdcrit}, then maximize retained pairs. Mean $|\mathrm{SMD}|$ is descriptive and does not override a failed feature-wise balance criterion.

\subsection{Stage 3: Motion--Quality Orthogonalization}

Matching controls the measured marginal quality variables under the stated balance criterion but does not guarantee balance of interactions, joint distributions, or feature-level dependence between $M$ and $Q$~\cite{resa2016matching}. We therefore perform a second, fold-wise sensitivity analysis. For each motion feature $M_j$, a linear nuisance model is fitted on the training fold:
\begin{align}
M_j &= g_j(Q)+\epsilon_j,\\
M_j^\perp &= M_j-\widehat{g}_j(Q).
\end{align}
The fitted transformation is applied unchanged to the test fold. This analysis tests whether discrimination survives removal of the linear component of motion predictable from measured quality. It is not presented as causal deconfounding.

\subsection{Stage 4: Robustness and Stress Tests}

The final stage evaluates whether the motion-versus-quality conclusion is stable across pair grouping, classifier family, motion representation, temporal aggregation, and abnormal category. These analyses are not intended to optimize predictive performance; rather, they stress-test whether the sign and magnitude of the controlled motion advantage depend on specific evaluation choices. Detailed protocols and results are reported in Sections 5 and 7.


\section{Experimental Protocol}
\label{sec:protocol}

\subsection{Dataset and Eligibility}

Our primary evaluation uses four human-centered UCF-Crime anomaly categories---Abuse, Assault, Fighting, and Shooting---against Normal. The candidate pool contains 50 videos from each anomaly category and 390 Normal videos. Consistent with the task definition in Sec. 3.1, the analysis is performed at the video level.

Eligibility for the observation-quality audit requires at least one retained person detection, because pose- and tracking-quality variables are otherwise undefined. After applying this criterion, the primary cohort contains 564 videos: 198 abnormal and 366 Normal. Category-level retention is 49/50 for Abuse, 50/50 for Assault, 50/50 for Fighting, 49/50 for Shooting, and 366/390 for Normal. Videos without a retained person detection are excluded rather than assigned artificial quality values. Accordingly, all primary conclusions are conditional on this eligible cohort.

\subsection{Observation and Motion Extraction}

A fixed YOLOv8-pose pipeline provides person bounding boxes, detection confidence, and pelvis/hip keypoints. A greedy nearest-neighbor tracker associates observations over time. Farneback optical flow is computed over the full frame and within tracked person regions. All videos are processed using the same frozen extraction settings.

We construct four base motion-derived representations: pelvis displacement, bounding-box-center displacement, person-localized flow, and whole-frame flow. We additionally evaluate Pelvis+BBox, which combines the two trajectory-based representations, and Multi-cue, which combines all four motion modalities. Each modality is summarized using video-level statistics that characterize the magnitude, variability, and temporal structure of the extracted motion signals.

Because both \(Q\) and the person-centered motion representations arise from the same detection--pose--tracking pipeline, we do not assume that they are statistically independent. Our audit instead asks whether motion-derived discrimination remains after explicit control of measured observation quality. Since Multi-cue also includes whole-frame flow, we interpret it as a combined motion-derived representation rather than a purely person-localized motion representation.

\subsection{Classifiers and Cross-Validation}

We use three fixed classifier families: $L_2$-regularized logistic regression ($C=1$), random forest (200 trees), and RBF-SVM ($C=1$, $\gamma=\texttt{scale}$). Hyperparameters are not tuned per representation.

The primary evaluation uses 5-fold stratified cross-validation repeated 10 times. Fold assignments are aligned across $Q$ and motion representations. Missing-value imputation and scaling are fit on the training fold only. Pair-aware sensitivity analysis uses matched-pair identifier as a group so the two members of a pair remain in the same fold.

ROC-AUC is the primary metric. PR-AUC, balanced accuracy, recall, and F1 are secondary metrics. Paired motion-minus-quality differences are summarized over the 50 aligned folds with 95\% paired resampling summary intervals. These intervals are used descriptively to summarize stability across aligned repeated-CV resamples and are not interpreted as independent-sample confidence intervals.

For temporal aggregation, we report the sign and magnitude of $\Delta$ across 36 prespecified combinations of window size, classifier, and aggregation rule.


\section{Main Results}
\label{sec:results}

\subsection{Observation Quality Is Already Highly Predictive}

The original cohort exhibits a strong observation-quality signal. The maximum pre-matching quality imbalance reaches $\max_k|\mathrm{SMD}_k|=0.712$. More importantly, this imbalance is predictive. Full $Q$ reaches ROC-AUC 0.827, 0.859, and 0.862 under LR, RF, and RBF-SVM, respectively. Multi-cue motion reaches 0.821, 0.844, and 0.827. Table \ref{tab:qdiagnostic} summarizes these results by quality-feature group.

\begin{table}[!t]
\centering
\small
\begin{tabular*}{\columnwidth}{@{\extracolsep{\fill}}lrrr@{}}
\toprule
Feature group & LR & RF & SVM \\
\midrule
Detection quality & 0.689 & 0.735 & 0.755 \\
Pose quality & 0.520 & 0.612 & 0.476 \\
Tracking quality & 0.819 & 0.814 & 0.830 \\
All $Q$ & \textbf{0.827} & \textbf{0.859} & \textbf{0.862} \\
Multi-cue motion & 0.821 & 0.844 & 0.827 \\
\bottomrule
\end{tabular*}
\caption{Original-cohort quality diagnostic ($n=564$). Observation quality alone matches or exceeds multi-cue motion performance across all three classifiers: Logistic Regression (LR), Random Forest (RF), and Support Vector Machine (SVM).}
\label{tab:qdiagnostic}
\end{table}

Tracking quality is the strongest individual quality group. Pose confidence alone is weak, while tracking continuity and fragmentation nearly reproduce the discrimination of the complete quality vector. In this cohort and fixed pipeline, observation reliability therefore provides a strong competing predictive explanation for part of the apparent discrimination.

\subsection{Cardinality Matching Achieves the Prespecified Marginal Balance Criterion}

Nearest-neighbor matching reduces average imbalance but fails the prespecified feature-wise criterion. At caliper 1.00 it retains 75 pairs, or 37.9\% of eligible abnormal videos, with mean $|\mathrm{SMD}|=0.065$ but maximum $|\mathrm{SMD}|=0.209$.

\begin{figure*}[htbp]
\centering
\begin{minipage}[t]{0.49\textwidth}
    \centering
    \includegraphics[
        width=\linewidth,
        trim=35 10 20 45,
        clip
    ]{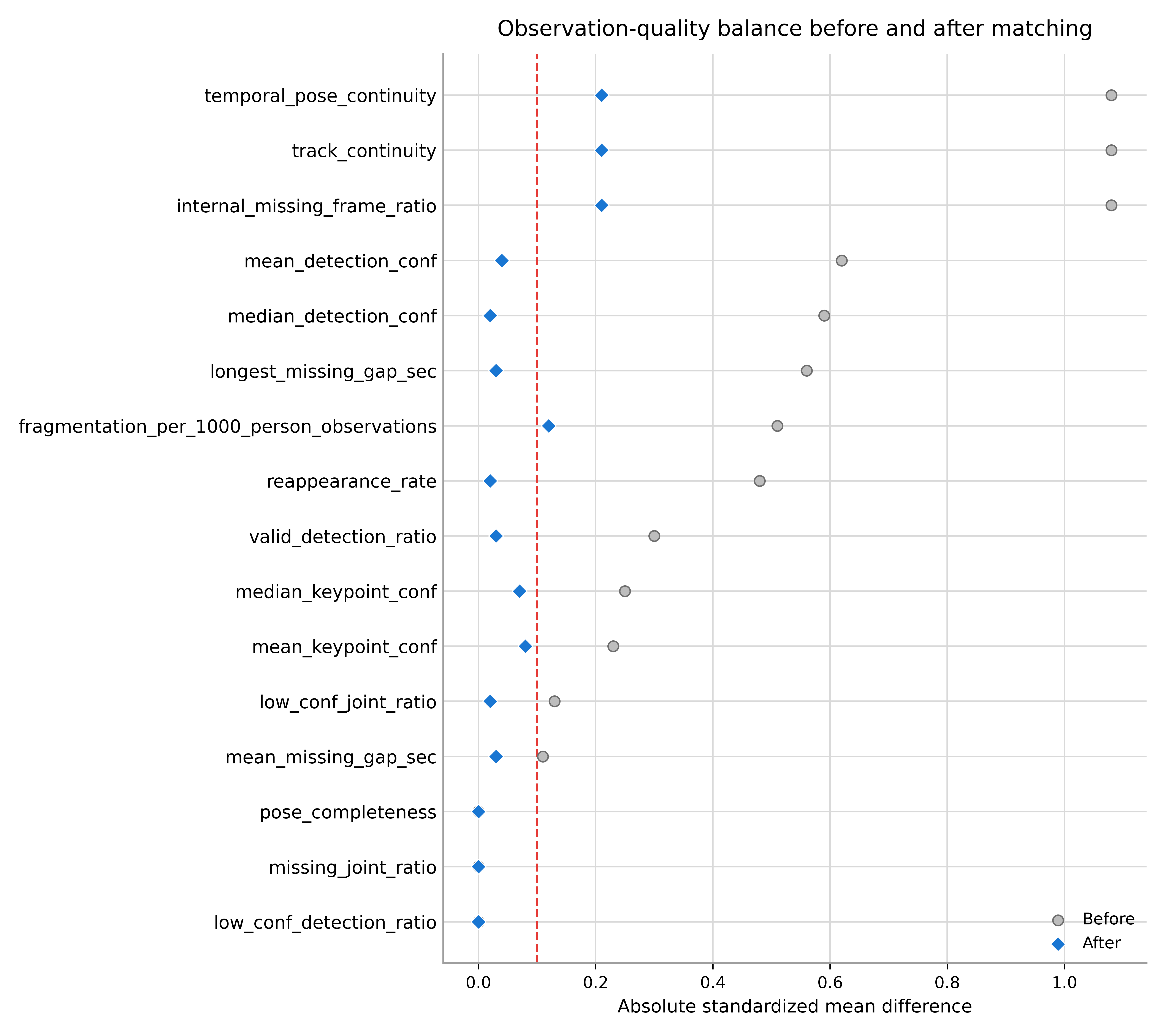}
    \vspace{0.8em}
    \textbf{(a)} NN-caliper 1.00
\end{minipage}
\hfill
\begin{minipage}[t]{0.49\textwidth}
    \centering
    \includegraphics[
        width=\linewidth,
        trim=35 10 20 45,
        clip
    ]{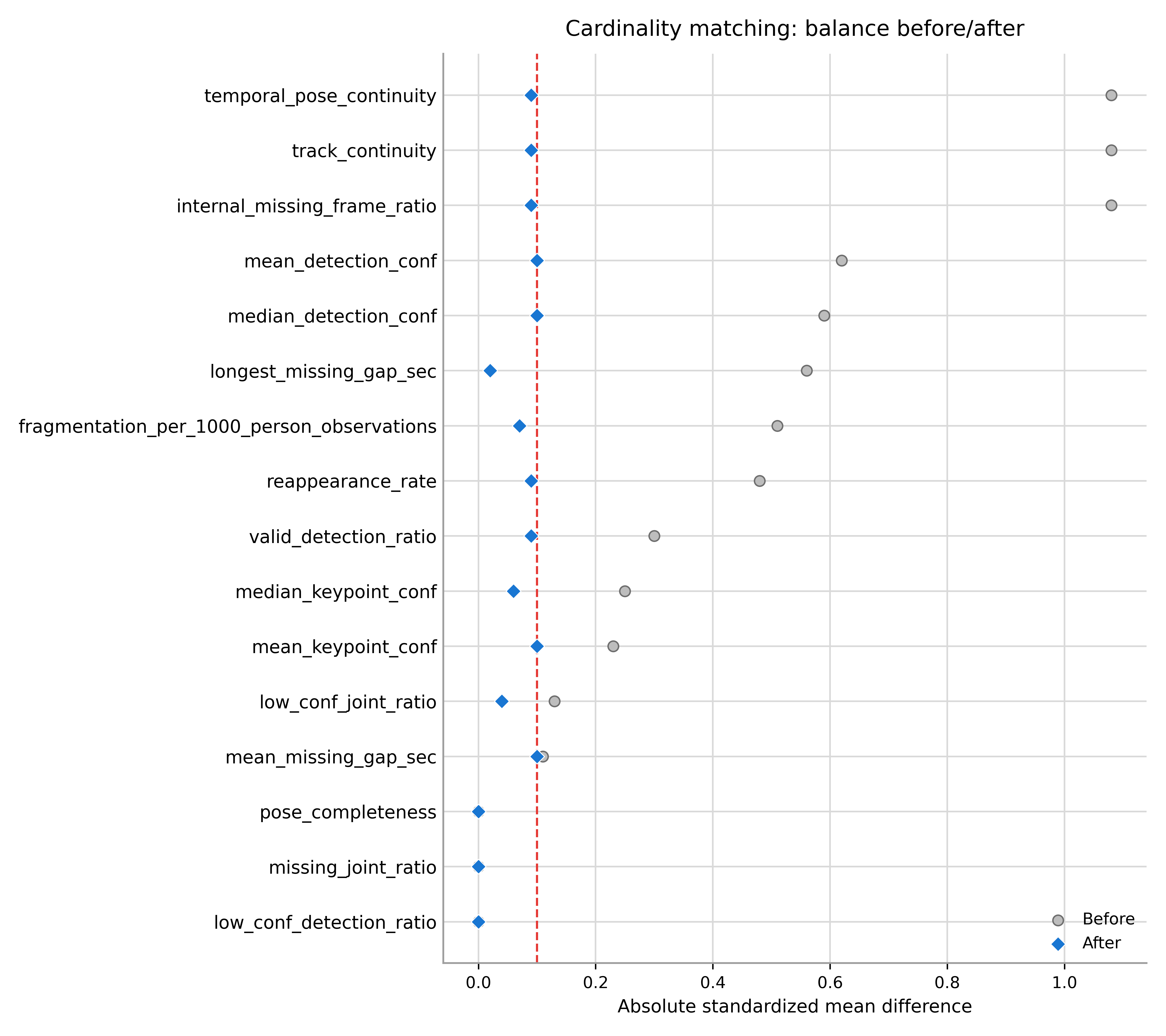}
    \vspace{0.8em}
    \textbf{(b)} Cardinality matching
\end{minipage}
\vspace{0.4em}
\caption{Observation-quality balance before and after matching. The dashed line marks $|\mathrm{SMD}|=0.10$. NN matching fails the criterion despite a smaller mean SMD; cardinality matching brings every measured quality variable below the threshold.}
\label{fig:matching_loveplots}
\end{figure*}

Cardinality matching retains 148 pairs (74.7\%) while satisfying the criterion for every active quality variable. This establishes feature-wise marginal mean balance, not equality of joint or nonlinear quality distributions: $\max|\mathrm{SMD}|=0.099$. We therefore use the 296-video cardinality cohort for the primary controlled comparison. Table \ref{tab:matching} compares the two matching strategies, and Figure \ref{fig:matching_loveplots} visualizes the resulting feature-wise balance.

\begin{table}[!t]
\centering
\small
\vspace{0.5em}
\begin{tabular*}{\columnwidth}{@{}l@{\extracolsep{\fill}}rrrr@{}}
\toprule
Method & Pairs & Ret. & Mean $|\mathrm{SMD}|$ & Max $|\mathrm{SMD}|$ \\
\midrule
NN-caliper 1.00 & 75 & 37.9\% & \textbf{0.065} & 0.209 \\
Cardinality & 148 & 74.7\% & 0.076 & \textbf{0.099} \\
\bottomrule
\end{tabular*}
\caption{Matching comparison. Only cardinality matching satisfies the feature-wise $|\mathrm{SMD}|<0.10$ requirement while retaining substantially more pairs.}
\label{tab:matching}
\end{table}

\subsection{After Measured-Quality Balancing, the Comparison Reverses}

The central result is the before--after reversal in motion versus quality. In the original cohort, multi-cue motion never exceeds full $Q$: $\Delta(M-Q)$ is $-0.006$ for LR, $-0.015$ for RF, and $-0.035$ for RBF-SVM. After cardinality matching, the differences become $+0.141$, $+0.046$, and $+0.065$, respectively. Table~\ref{tab:mainmatrix} reports the complete original, matched, and orthogonalized matrix.

Aligned repeated-CV resamples show the same directional reversal. In the original cohort, LR is indistinguishable from zero ($-0.006$; 95\% interval $[-0.019,0.006]$), while RF and RBF-SVM favor quality. In the matched cohort, all three intervals favor motion: LR $+0.141$ $[0.113,0.168]$, RF $+0.046$ $[0.023,0.070]$, and RBF-SVM $+0.065$ $[0.042,0.088]$.

This reversal is the key evaluation finding, but not evidence that quality has been removed. After matching, quality-only RF still reaches 0.732 AUC. SMD constrains feature-wise means, whereas a flexible classifier exploits nonlinear or joint structure~\cite{resa2016matching}. We therefore treat post-match quality-only AUC as a required balance diagnostic alongside the Love plot: the controlled result supports an additional multi-cue advantage, not quality-free motion.

\subsection{The Effect Is Representation-Dependent}

The reversal is not shared equally by every motion representation. Single modalities vary across classifiers, whereas multi-cue motion is consistently strongest. These 18-statistic-per-cue descriptors are handcrafted: they test whether explicit measured motion survives quality control, not whether learned pose embeddings or stronger VAD systems do. Learned-model generalization remains an empirical requirement.

Because the combined representation includes whole-frame flow, this result should not be interpreted as evidence that all of the surviving signal is person-localized. Rather, it shows that the combined motion-derived representation retains discrimination beyond the measured quality baseline.

\begin{table*}[!t]
\centering
\small
\vspace{0.5em}
\begin{tabular*}{\textwidth}{@{}l@{\extracolsep{\fill}}rrrrrrrrr@{}}
\toprule
& \multicolumn{3}{c}{Original AUC} &
\multicolumn{3}{c}{Matched AUC} &
\multicolumn{3}{c}{Orthogonalized AUC} \\
\cmidrule(lr){2-4} \cmidrule(lr){5-7} \cmidrule(lr){8-10}
Representation & LR & RF & SVM & LR & RF & SVM & LR & RF & SVM \\
\midrule
Quality only & 0.827 & 0.859 & 0.862 & 0.586 & 0.732 & 0.681 & -- & -- & -- \\
Pelvis & 0.774 & 0.802 & 0.783 & 0.642 & 0.711 & 0.655 & 0.674 & 0.716 & 0.711 \\
BBox center & 0.778 & 0.817 & 0.786 & 0.669 & 0.725 & 0.690 & 0.710 & 0.749 & 0.761 \\
Person flow & 0.667 & 0.765 & 0.667 & 0.621 & 0.732 & 0.675 & 0.710 & 0.720 & 0.727 \\
Whole flow & 0.744 & 0.769 & 0.733 & 0.662 & 0.701 & 0.623 & 0.652 & 0.631 & 0.617 \\
Multi-cue & 0.821 & 0.844 & 0.827 & \textbf{0.726} & \textbf{0.778} & \textbf{0.746} & \textbf{0.752} & \textbf{0.788} & \textbf{0.783} \\
\bottomrule
\end{tabular*}
\caption{Main experimental matrix. The controlled result is strongest and most stable for the combined motion representation.}
\label{tab:mainmatrix}
\end{table*}


\section{Robustness and Stress Tests}
\label{sec:stress}

\subsection{Pair-Aware Cross-Validation}

The primary repeated-stratified CV sometimes splits the two members of a matched pair across folds. We therefore repeat the analysis with matched-pair identifier as a grouping variable. The result does not change direction. Multi-cue motion exceeds quality-only by $+0.115$, $+0.039$, and $+0.043$ ROC-AUC for LR, RF, and RBF-SVM, respectively (Table \ref{tab:pairaware}). All 15 classifier–metric point differences remain positive, and the primary ROC-AUC comparison favors motion under all three classifiers.

\begin{table}[!t]
\centering
\small
\vspace{0.5em}
\begin{tabular*}{\columnwidth}{@{}l@{\extracolsep{\fill}}rrr@{}}
\toprule
Model & Pair-aware $Q$ & Pair-aware Motion & $\Delta$ \\
\midrule
LR & 0.609 & 0.724 & $+0.115$ \\
RF & 0.730 & 0.769 & $+0.039$ \\
RBF-SVM & 0.715 & 0.758 & $+0.043$ \\
\bottomrule
\end{tabular*}
\caption{Pair-aware ROC-AUC on the cardinality-matched cohort.}
\label{tab:pairaware}
\end{table}

\subsection{Motion--Quality Orthogonalization}

Linear residualization also preserves the multi-cue result. Its AUC increases from 0.726 to 0.752 under LR, 0.778 to 0.788 under RF, and 0.746 to 0.783 under RBF-SVM. In contrast, whole-frame optical flow decreases under every classifier, most sharply under RF (0.701 to 0.631).

This contrast is informative. The full multi-cue representation retains discrimination after removing the linear component predictable from $Q$, whereas whole-frame flow loses discrimination, indicating that a larger fraction of its predictive signal is linearly predictable from measured Q in this analysis.

\subsection{Temporal Aggregation}

We recompute motion summaries with 1, 2, 4, and 8-second windows and aggregate window scores by mean, maximum, or top-$k$ mean. The motion advantage remains positive in 32 of 36 prespecified settings, with mean $\Delta\text{AUC}$ ranging from $+0.076$ to $+0.094$ across window sizes (Table \ref{tab:temporal}).

\begin{table}[!t]
\centering
\small
\begin{tabular*}{\columnwidth}{@{}l@{\extracolsep{\fill}}rrr@{}}
\toprule
Window & Mean $\Delta\text{AUC}$ & Positive $\Delta\text{AUC}$ & Recurring exception \\
\midrule
1 s & +0.081 & 8/9 & SVM + mean \\
2 s & +0.090 & 8/9 & SVM + mean \\
4 s & +0.094 & 8/9 & SVM + mean \\
8 s & +0.076 & 8/9 & SVM + mean \\
\bottomrule
\end{tabular*}
\caption{Temporal aggregation sensitivity for multi-cue motion.}
\label{tab:temporal}
\end{table}

The conclusion is therefore stable across temporal scales but not invariant to every downstream decision rule. RBF-SVM with mean aggregation is negative at all four windows, providing a consistent counterexample rather than an isolated fluctuation.

\subsection{Exploratory Category Dependence}

Category-level patterns are heterogeneous \ref{tab:category}: Fighting shows a large raw motion advantage, whereas Assault and Shooting favor quality before orthogonalization. Because category-wise sample sizes are limited, we treat these results as descriptive stress tests; the pooled advantage does not imply a uniform violence signature.

\begin{table}[!t]
\centering
\small
\label{tab:category}
\vspace{0.5em}
\begin{tabular*}{\columnwidth}{@{}l@{\extracolsep{\fill}}rrrrr@{}}
\toprule
Category & $Q$ & Raw M & Orth. M & Raw$-Q$ & Orth.$-Q$ \\
\midrule
Abuse & 0.544 & 0.586 & 0.687 & +0.042 & +0.143 \\
Assault & 0.723 & 0.673 & 0.641 & -0.050 & -0.082 \\
Fighting & 0.622 & 0.833 & 0.745 & +0.211 & +0.123 \\
Shooting & 0.574 & 0.521 & 0.641 & -0.053 & +0.067 \\
\bottomrule
\end{tabular*}
\caption{Category-wise multi-cue comparison. Aggregate motion discrimination is not a single uniform violence signature.}
\label{tab:category}
\end{table}

This result prevents an overly broad conclusion. The matched multi-cue advantage is robust at the pooled task level, but it does not mean that every violence category is separated by the same motion evidence.


\section{Discussion}
\label{sec:discussion}

\subsection{What the Audit Establishes}

The strongest result of this study is not a small change in AUC; it is a change in what the AUC means. In the original cohort, detection--pose--tracking reliability alone is as discriminative as multi-cue motion. A person-centered VAD result measured only on that distribution does not uniquely demonstrate abnormal-motion modeling.

The matched analysis reduces part of this ambiguity. Once every measured quality variable is brought below the prespecified marginal-balance threshold, the multi-cue representation exceeds the quality-only baseline under all three classifier families. Pair-aware CV and linear orthogonalization preserve the same direction. The data therefore support two statements simultaneously:
\begin{enumerate}[leftmargin=*]
    \item UCF-Crime contains a strong observation-quality shortcut for this person-observation pipeline.
    \item Measured observation quality does not account for the full multi-cue motion discrimination after balancing.
\end{enumerate}
Neither statement requires a causal claim.

\subsection{Residual Dependence After Quality Balancing}

Observation quality and motion are physically coupled because both arise from the same video and, for person-centered modalities, the same detections and tracks. Our objective is not to prove statistical independence between $Q$ and $M$. The matched comparison therefore asks a narrower question: whether the motion-derived representation retains incremental discrimination under explicit control of the measured marginal quality variables.

This distinction also explains the residual quality-only RF AUC of 0.732 after matching. The SMD criterion balances marginal means; it does not force equality of nonlinear or joint quality distributions. That residual predictability is therefore an observed limitation of the balancing procedure, not a reason to discard the matched comparison. It is precisely why we report quality-only performance after matching rather than treating a Love plot as proof that quality has disappeared.

\subsection{Implications for Person-Centered VAD Evaluation}

We recommend three complementary checks for person-centered VAD: (i) quality-only prediction from detection–pose–tracking reliability, (ii) explicit balance diagnostics for measured quality, and (iii) post-balance comparison of motion against the remaining quality-only baseline under identical evaluation conditions. In our data, these checks change the interpretation of the benchmark while category heterogeneity limits any claim of universal violence understanding.


\section{Limitations}
\label{sec:limitations}
Our study has four main limitations.

First, measured observation quality is only a proxy for observability. The quality vector Q summarizes detection, pose, and tracking reliability from a fixed pipeline, but does not directly measure factors such as true keypoint error, identity switches, occlusion, camera motion, subject scale, or compression.

Second, cardinality matching achieves marginal rather than complete balance. Although all measured quality variables satisfy $\max_k |\mathrm{SMD}_k| < 0.10$, nonlinear interactions and joint distributions may remain imbalanced. The post-match quality-only RF AUC of $0.732$ confirms residual predictability. We therefore interpret matching as control of measured marginal quality, not removal of all quality-related information.

Third, observation quality and motion-derived representations are not fully separable. Both Q and motion features originate from the same detection--pose--tracking pipeline, and some motion summaries depend on valid observations. Matching and fold-wise orthogonalization therefore test whether motion retains incremental discrimination beyond measured Q, rather than establishing a pure quality--motion decomposition.

Fourth, empirical generality remains limited. The primary analysis covers four human-centered UCF-Crime categories, one fixed observation pipeline, and handcrafted motion-derived representations, while category-wise results show substantial heterogeneity. The matched conclusions therefore apply to the evaluated cohort and pipeline rather than establishing a category-invariant or dataset-general property of person-centered VAD. Generalization across datasets, observation pipelines, and learned VAD representations remains to be established.


\section{Conclusion}
\label{sec:Conclusion}
We audited whether person-centered VAD discrimination can be explained by the reliability of the observation pipeline from which motion is constructed. On the original UCF-Crime cohort, measured detection–pose–tracking quality is as predictive as multi-cue motion. After cardinality matching achieves the prespecified marginal balance criterion, the combined motion-derived representation consistently exceeds the remaining quality-only baseline across three classifier families, with the same direction under pair-aware validation and fold-wise orthogonalization. These results do not establish quality-free or causal motion understanding; rather, they show that VAD performance becomes more interpretable when observation reliability is explicitly audited and controlled. We therefore advocate observation-quality diagnostics as a routine complement to person-centered VAD evaluation.


{
    \small
    \bibliographystyle{ieeenat_fullname}
    \bibliography{main}
}

\end{document}